\documentclass{article}

\usepackage{arxiv}

\usepackage[utf8]{inputenc} % allow utf-8 input
\usepackage[T1]{fontenc}    % use 8-bit T1 fonts
\usepackage{hyperref}       % hyperlinks
\usepackage{url}            % simple URL typesetting
\usepackage{booktabs}       % professional-quality tables
\usepackage{amsfonts}       % blackboard math symbols
\usepackage{nicefrac}       % compact symbols for 1/2, etc.
\usepackage{microtype}      % microtypography
\usepackage{graphicx}
\usepackage{natbib}
\usepackage{doi}
\usepackage{amsmath}
\usepackage{array}
\usepackage{tabu}	% needed for kable
\usepackage{multirow}

\title{Semiparametric Latent Topic Modeling on Consumer-Generated Corpora}
\date{}

\author{ \hspace{1mm}Dayta, Dominic B.\\
	School of Statistics\\
	University of the Philippines\\
	Diliman, Quezon City \\
	\texttt{dbdayta@up.edu.ph} \\
	\And
	\hspace{1mm}Barrios, Erniel B. \\
	School of Statistics \\
	University of the Philippines\\
	Diliman, Quezon City \\
	\texttt{ebbarios@up.edu.ph} \\
}

\renewcommand{\headeright}{}
\renewcommand{\undertitle}{}
\renewcommand{\shorttitle}{Semiparametric Topic Modeling}

\hypersetup{
pdftitle={Semiparametric Latent Topic Modeling on Consumer-Generated Corpora},
pdfauthor={Dominic B.~Dayta, Erniel B.~Barrios},
pdfkeywords={Topic Modelling, Semiparametric Regression, Latent Dirichlet Allocation, Nonnegative Matrix Factorization},
}

\begin{document}
\maketitle

\begin{abstract}
Legacy procedures for topic modelling have generally suffered problems of overfitting and a weakness towards reconstructing sparse topic structures. This paper proposes SemiparTM, a two-step approach utilizing nonnegative matrix factorization and semiparametric regression in topic modeling. SemiparTM enables the reconstruction of sparse topic structures in the corpus and provides a generative model for predicting topics in new documents entering the corpus. Assuming the presence of auxiliary information related to the topics, this approach exhibits better performance in discovering underlying topic structures in cases of corpora that are small and limited in vocabulary. In an actual consumer feedback corpus, SemiparTM also demonstrably provides interpretable and useful topic definitions comparable with those produced by the legacy methods.
\end{abstract}


\section{Introduction}
\label{sec:section1}

The fields of natural language processing and information retrieval saw a productive past two decades due largely to the emergence and worldwide adoption of two modern technologies: large-scale document indexing and storage facilities, of which perhaps the two most prominent brands are JSTOR and Google Books, and social networking sites that allow individual users to create and distribute various types of content, a considerable fraction of which exist in the form of texts (status updates, blog posts, and \textit{tweets}). All these have led to a relentless growth in information-rich but unstructured collections of text data – referred to as \textit{corpora} in natural language terminology – in terms of volume, velocity, and frequency such that manual approaches to document indexing and classification are quickly becoming obsolete.

Outside the context of online archives, methods that enable automated classification and analysis of voluminous corpora would prove to be valuable technology. It has been applied to legal research \citep{ravi2012} and for analyzing patterns behind railroad accidents \citep{williams2018comparison}. In the commercial space, companies can take advantage of thousands of posts being contributed by users on a daily basis about their products and services on social media and review aggregator websites like \textit{Yelp} and \textit{TripAdvisor}.

Among the core functions of Customer Relations Management (CRM) departments in customer-facing industries is capturing what they call the Voice of the Customers (VOC). VOC refers to feedback, self-reported by the customers in the form of verbatim complaints, comments, inquiries, and the likes sent in via one or more points of capture. Presently, the standard approach that industries have taken towards the capture and analysis of VOC is via the employment of a Business Process Outsourcing (BPO) partner, that would, in turn, deploy customer care representatives to receive and process feedback. Representatives are trained in handling customers and are oriented towards categorizing feedback according to subject. Through this arrangement, previously unstructured data from call transcripts, e-mails, SMS, social media posts, and other possible venues are transformed into structured (i.e., tabular) summaries which are then sent to the client company for resolution, actions, and further analysis.

The proliferation of social networking and micro-blogging services on the Internet has given consumers an inexhaustible variety of platforms through which they may voice out satisfaction or dissatisfaction towards these companies’ products and services. All these have led to a relentless growth for VOC in terms of volume, velocity, and frequency such that manual approaches to feedback management are quickly becoming inefficient. Successful formulation of a new methodology for automated complaints classification would not only impact businesses directly concerned, but also their outsourced service providers as this would ease the growing tedium of manual feedback capture systems and allow for better, more strategic allocation and management of manpower.

This need for automation is hardly novel in the literature. Hotel reviews on certain travel websites have been analyzed to the effect of identifying driving factors to customer satisfaction \citep{berezina2016understanding}. This was accomplished by grouping together known words appearing in the reviews under general tokens that identify thematic similarities between customers' complaints and commendations. Other, more sophisticated approaches involve the use of fuzzy algorithms \citep{razali2016classification}.

Both approaches can be seen as forerunners to the use topic modeling for analyzing VOC, wherein the topic structures were defined \textit{a priori} by the researchers (or, in the latter case, through fuzzy logic). In true topic modeling, these structures are \textit{discovered} rather than pre-determined by the analyst, and this discovery provides the objective of the algorithm. In \citet{coussement2008improving}, Latent Semantic Analysis (LSA) is used to extract linguistic characteristics from customer complaints, and these characteristics were later used as features in a classification model.

The method of Singular Value Decomposition (SVD) is used to discover underlying ``semantic structures" defined by word co-occurrences, \citet{deerwester1988improvinginformation}. This method, like the others that would succeed it, depended on a specific representation of the corpora into a matrix form that is much more suited for statistical analysis. By representing each document as a vector defined by its frequencies across a set of unique words, the document vectors together formed a matrix for the entire corpus, which could be subjected to factorization via SVD. This would be refined with Probabilistic Latent Semantic Analysis or PLSA \citep{hoffmanplsa} which provided a more interpretable framework by defining the topics as probability distributions over words, and replacing SVD with a more formal estimation procedure via the Expectation-Maximization (EM) algorithm. Later, Latent Dirichlet Allocation(LDA) addressed some of PLSA’s shortcomings by giving it a Bayesian flavor \citep{blei2003latent}. Nevertheless, LSA has arguably set the direction of much of the research regarding topic modeling.

This paper proposes a two-stage semiparametric topic modeling (SemiparTM) approach. The first stage is based on LSA, replacing the use of SVD with nonnegative matrix factorization with the goal of addressing LSA’s lack of clear interpretations towards topic distributions. A second stage that makes use of semiparametric regression is added to address criticisms that have also been levelled against PLSA and LDA. Specifically, the semiparametric method is aimed at accurately reproducing latent topic structures in the corpus, providing a generative model that yields better prediction of topics in new documents that have yet to enter the corpus.

The rest of this paper is organized as follows: in the next section, we formalize the vector space representation for converting unstructured text documents into a numerical format on which we can characterize statistically. This section also covers some basic notations that will be useful for discussing the formulation of SemiparTM in the Section \ref{sec:section3}. Section \ref{sec:section4} outlines the simulation study by which we intend to evaluate and compare the performance of SemiparTM against LSA, PLSA, and LDA in discovering latent topic structures, the results of which are then presented in Section \ref{sec:section5}. Application of the proposed method on actual consumer feedback is presented in Section \ref{sec:section6}. Finally, we close our discussion with some concluding remarks and practical recommendations for using SemiparTM on real-world corpora.

\section{Topic Modeling}
\label{sec:section2}

Topic modeling requires a way of translating unstructured texts into quantities on which statistical models can be trained. One popular approach used in the legacy methods as well as a number of other succeeding methods is the \textit{Vector Space Model}, or the \textit{Bag of Words} approach. In this approach, documents are converted into vectors whose components correspond to scores for unique words existing in the entire corpus. Note that \textit{score} is used as a general term towards a variety of procedures to representing, in numerical form, the level of occurrence, importance, or otherwise of words in the documents of a corpus.

\begin{figure}
	\centering
	\includegraphics[scale=0.5]{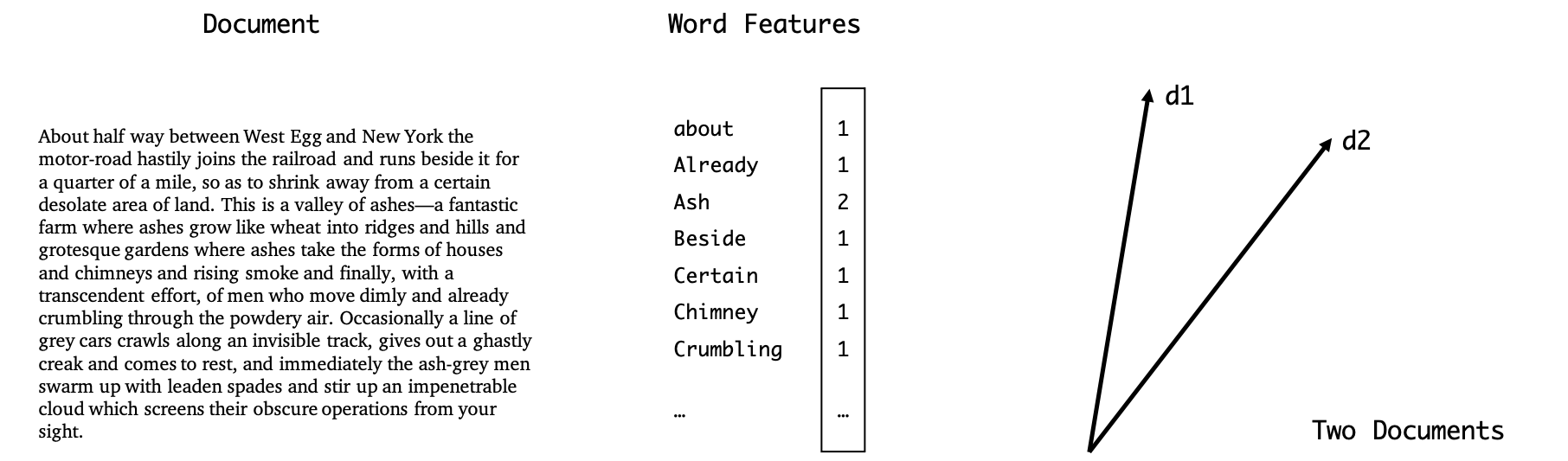}
	\caption{The Vector Space Model}
	\label{fig:vsm}
\end{figure}

Suppose a corpus is composed of $D$ documents which together make use of a total of $W$ unique words. A document $d_i$ in the corpus may be referred to as the $W \times 1$ vector of scores, $d_i  = [ w_1i,w_2i,w_3i,...,w_Wi ]$. This process of translating documents into vectors is illustrated in Figure \ref{fig:vsm}. All the $D$ documents in the corpus can then be stacked into a $W \times D$ matrix $\mathbf{Y}$ where:

\begin{equation}
\label{eq:vsm}
\mathbf{Y}_{W \times D} = 
\begin{bmatrix}
w_{11} & w_{12} & w_{13} & ... & w_{1D} \\
w_{21} & w_{22} & w_{23} & ... & w_{2D} \\
... & ... & ... & ... & ... \\
w_{W1} & w_{W2} & w_{W3} & ... & w_{WD}
\end{bmatrix} = [ d_1, d_2, d_3,  ..., d_D ]
\end{equation}

In this formulation, the values $Y_ij$ in matrix $\mathbf{Y}$ refer to the score for word $i$ in document $j$ \citep{zhu2012sparse}.

Under LSA, topics are latent variables that manifest in the co-occurrence of particular words in a document. These latent variables are discovered with the use of SVD, decomposing the corpus matrix as $\mathbf{Y} = \mathbf{X} \mathbf{S} \mathbf{B}$, where $\mathbf{X}$, $\mathbf{S}$, and $\mathbf{B}$ are $W \times T$, $T \times T$, and $T \times D$, respectively \citep{deerwester1988improvinginformation}. $\mathbf{X}$ and $\mathbf{B}$ are both orthogonal matrices, and $\mathbf{S}$ is a diagonal matrix containing singular values for $\mathbf{B}$. SVD maps the original word-document scores in matrix $\mathbf{Y}$ onto a lower-dimensional latent space, such that the documents are represented by a set of linearly independent base vectors \citep{steinberger2004using}.

$\mathbf{X}$ is interpreted as a \textit{dictionary} matrix, where $X_{ik}$ gives the level of importance of word $i$ to topic $k$. Meanwhile, $\mathbf{B}$ is interpreted as the \textit{topic distribution matrix}, with $B_{kj}$ gives the level of expression of topic $k$ in document $j$. A word that has a stronger association to a certain topic will have a higher value in $\mathbf{X}$. At the same time, a document that contains more of words related to a certain topic will have a higher value in $\mathbf{B}$.

PLSA takes a more probabilistic approach to topic modeling \citep{hoffmanplsa}. Letting $d$ and $w$ be indices referring, respectively, to a specific document and word, while $z$ as an index for topic, the joint probability of observing a word and a document is given by $P(d,w)$ as

\begin{align}
\label{eq:plsa}
P(d,w) &= P(d) \sum_{t \in \{ 1,2,...,T\}}{P(w|t) P(t|d)} = \sum_{t \in \{ 1,2,...,T\}}P(t)P(d|t)P(w|t)
\end{align}

The probability model in \ref{eq:plsa} appears analogous to LSA. $P(d,w)$ can be viewed as a $D \times W$ matrix similar to the matrix $\mathbf{Y}$ of word scores, while $P(t)$, $P(d|t)$, and $P(w|t)$ are analogous to matrices $\mathbf{S}$, $\mathbf{B}$, and $\mathbf{X}$. Instead of using a matrix decomposition procedure like SVD, PLSA is estimated using the EM algorithm, see for example \citet{hoffmanplsa} for further details.

PLSA, however, lacks a probabilistic model at the level of documents \citep{blei2003latent}. There is proper model for what produces $P(t|d)$ in equation \ref{eq:plsa}. Treating d as an index, this implies that the number of parameters to be estimated grows linearly with the documents in the corpus. At the same time, supposing there are $d \in \{1,2,...,D\}$ documents in the corpus at training time, the model will have no way of modeling any further, yet-unseen documents at testing time, since any indices $d \in \{D+1,D+2,...\}$ will have been outside the bounds of the model.

LDA corrects this problem with redefining topic expressions as arising from a Dirichlet prior \citep{blei2003latent}. A document is re-expressed as a vector over words, $w = \{w_1,w_2,…,w_{Wd}\}$  where $W_d$ is the number of words in the sequence defining the $d$th document. The generative model for this sequence is as follows:

\begin{itemize}
	\item $W_d \sim Poisson(\eta)$ 
	\item $\mathbf{\theta} \sim Dirichlet(\alpha)$
	\item For each of the $W_d$ words, $w_j$
	\begin{itemize}
		\item Choose topic $Z_k \sim Multinomial(\mathbf{\theta})$
		\item Choose a word $w_j$ from $P(w_j | z_k, \beta)$, multinomial
	\end{itemize}
\end{itemize}

Given the parameters$\alpha$ and $\beta$, inference can then be done using the joint distribution

\begin{equation}
\label{eq:lda}
P(\mathbf{\theta},\mathbf{z},\mathbf{w} | \alpha, \beta) = P(\mathbf{\theta}|\alpha) \prod_{k=1}^{L}{P(z_k|\mathbf{\theta})P(w_k|z_k,\beta)}
\end{equation}

The distribution $\mathbf{\theta}$ already refers to the vector of topic expressions per document and is analogous to the $\mathbf{X}$ matrix in LSA. LDA's advantage to PLSA is given by the fact that in this formulation, the topic distribution $\mathbf{\theta}$ unlike $\mathbf{X}$ is trained once from the data. After the parameters of the Dirichlet distribution have already been trained, and the posterior probability already suitably estimated, LDA faces no problems in evaluating new documents \citep{blei2003latent}. In the event that the trained model encounters a document previously unseen, it simply generates a random sample from the trained posterior \citep{srivastava2009text}.

\citet{zhu2012sparse} argue that this particular approach cannot properly take into account the sparsity that ought to be present in realistic settings due to the normed nature of its output (topic expressions are probabilities that must sum to 1). Other criticisms leveled at LDA include its instability and issues with replicability \citep{agrawal2018wrong}. Researchers have found issues reproducing its results \citep{agrawal2018wrong,barua2014developers} and in certain cases LDA seems to offer little to no improvement to either PLSA or LSA in terms of empirical performance on real data. LSA has been demonstrated to outperformed both LDA and PLSA in terms of accuracy \citep{krestel2009latent} especially on small corpora (150 documents), as \citet{kakkonen2008comparison} and \citet{blei2003latent} both recommended the use of 1,000 to 3,000 documents for training LDA.

Despite continuing widespread adoption of LDA in topic modeling research, the literature has since progressed to more advanced methodologies that address these known limitations in legacy topic models. An entire class of these new methods are motivated with replacing the standard vector space model, which treat words as atomic units, or exchangeable vector components, and in its place employing word embeddings. Word embeddings represent the words as vectors whose distance can be measured in reference to other word vectors \citep{mikolov2013efficient}. In consequence, a word that is contextually similar (e.g., ``food" and ``eat") or syntactically related (``thank" and ``you") appear with much more proximity in the vector space. Word embeddings have been shown to perform better in assessing syntax and semantic relationship between words in corpora in comparison to other neural network based methods.

An LDA implementation built on top of word embeddings (lda2vec), has been demonstrated to produce topics that have much better syntactic and semantic cohesion, as contextual relationship between words have been preserved with word embeddings \citep{moody2016mixing}. A similar reformulation of LDA based on word embeddings involves adding multivariate Gaussian distributions over the embedding space and estimated using a collapsed Gibbs sampler. This Gaussian LDA is found to produce similar topics as a standard LDA, but has the advantage of properly accounting for out-of-vocabulary words: words that were not part of the corpus at training time, but encountered in the hold-out set \citep{dasetal2015gaussian}. Finally, a nonparametric method of word-embeddings-based topic modeling that applies a combination of hierarchical Dirichlet processes while regularizing word densities into a unit sphere via the von Mises-Fisher distribution has provided a way by which the number of existing topics can be flexibly discovered in an unsupervised manner \citep{BatmanghelichSN16}.

\section{Semiparametric Topic Modeling (SemiparTM)}
\label{sec:section3}

This section presents detailed formulation of SemiparTM. We introduce the two stages of the method, along with the cross-validation process for tuning its shrinkage penalty.

Consider the corpus we formulated in \ref{eq:vsm} containing raw word counts. Suppose that the number of topics expected in the corpus, $T$, is known beforehand. We wish to decompose matrix $\mathbf{Y}$ into two matrices $\mathbf{X}$ and $\mathbf{B}$, correspondingly of dimensions $W \times T$ and $T \times D$, where the former is a matrix of topic distribution over words (how likely a word is to appear under a certain topic) and the latter is a matrix of topic expression over documents (how much of a topic is present in a document). Factorization of $\mathbf{Y}$ then becomes a matter of finding appropriate matrices $\mathbf{X}$ and $\mathbf{B}$ both non-negative that minimizes the objective function

\begin{equation}
    \label{eq:nmf}
    (\hat{\mathbf{X}},\hat{\mathbf{B}}) = \arg \min_{(\mathbf{X},\mathbf{B})}{||\mathbf{Y} - \mathbf{X} \mathbf{B}||^2_2 + \xi (||\mathbf{X}||_1 + ||\mathbf{B}||_1)}
\end{equation}

where $||.||_2^2$ represents the $L_2$ norm while $||.||_1$ represents the $L_1$ norm. Minimization of \ref{eq:nmf} is done through a similar gradient descent algorithm according to \citep{Liu2010}, with multiplicative update rules presented in equations \ref{eq:nmf-step1} and \ref{eq:nmf-step2}.

\begin{align}
    \mathbf{X}_{ik} & = \mathbf{X}_{ik}
    \frac{(\mathbf{Y} \mathbf{B}^T)_{ik}}{(\mathbf{X} \mathbf{B} \mathbf{B}^T)_{ik}+\xi} \label{eq:nmf-step1} \\
    \mathbf{B}_{kj} & = \mathbf{B}_{kj}
    \frac{(\mathbf{X} \mathbf{Y})_{kj}}{(\mathbf{X}^T \mathbf{X} \mathbf{B})_{kj}+\xi} \label{eq:nmf-step2}
\end{align}

The constant $\xi \geq 0$ is a tuning parameter controlling the level of sparsity in the estimated matrices $\mathbf{X}$ and $\mathbf{B}$. Higher values of $\xi$ would lead to more components of the two matrices being populated by zeroes. We simplify this parameter to apply to both $\mathbf{X}$ and $\mathbf{B}$, though the optimization may also be performed giving different sparsity penalties for either (e.g., $\xi_1$ for $\mathbf{X}$, $\xi_2$ for $\mathbf{B}$). The sparsity penalty is supposed to represent what \citet{zhu2012sparse} argue must be the natural behavior of topic distributions: that topics are defined by the co-occurrence of very few, specific words, while documents can only have a very narrow set of topics. The value of $\xi$ can be set to any fixed nonnegative constant, or optimized via cross-validation as follows:

\begin{itemize}
	\item Divide the matrix $\mathbf{Y}$ randomly into $K$ equal folds, such that $\mathbf{Y}^{(1)},\mathbf{Y}^{(2)},…,\mathbf{Y}^{(K)}$ form partitions of $\mathbf{Y}$
	\item For each $k=1,2,…,K$ take the fold $\mathbf{Y}^{(k)}$ as the training set $\mathbf{Y}_{train}$. Combine the rest of the $K-1$ folds to form $\mathbf{Y}_{test}$.
	\begin{itemize}
		\item Perform nonnegative LSA on $\mathbf{Y}_{train}$ to obtain $\mathbf{X}_{train}$ and $\mathbf{B}_{train}$
		\item Estimate $\mathbf{B}_{test}$ via semiparametric regression stage on $\mathbf{Z}_{test}$
		\item Measure resulting error $E_k = || \mathbf{Y}_{test} - \mathbf{X}_{train} \mathbf{B}_{test} ||_2^2$
		\item Take $\xi = \arg \min_{k}{E_k}$
	\end{itemize}
\end{itemize}

Stage 2: Regression. Aside from a word frequency matrix $\mathbf{Y}$, we also assume that there exists with the corpus a data matrix $\mathbf{Z}$ containing external information regarding the documents themselves. Suppose there are $p$ of these external variables $\mathbf{Z}_1,\mathbf{Z}_2,...,\mathbf{Z}_p$. The second stage of the topic modeling procedure models the topic distributions of the documents against these external variables using semiparametric regression, with the topic expressions as dependent variables.

For a set of $D$ documents, a single row of the matrix $\mathbf{B}$, denoted as $b^{(t)}$ where $t$ is any one of the $T$ topics generated, a hierarchical equation can be placed over $b^{(t)}$ such that for the $i$th document:

\begin{equation}
    \label{eq:semipar1}
    b^{(k)}_d = \beta^{(k)}_0 + \sum_{l=1}^{p}{f^{(k)}_l (Z_{dl})}+\epsilon_{dl}
\end{equation}

where $\varepsilon_{il} \sim Normal(0,1)$ and each $f_l^{(t)}$ is some function describing the relationship of external variable $Z_l$ with the topic $t$ as represented by the values of $b^{(t)}$. The semiparametric method allows using any regression method to estimate $f_l^{(t)}$, however, for the purposes of this paper, we use B-splines.

What follows is a brief note on the conceptual justification for this regression stage. PLSA is over-parametrized since it has no generative model for the distribution of topics in a particular document \citep{blei2003latent}. The regression stage resolves this by substituting $P(t | d)$ in PLSA for a function that provides the values for topic expressions $t$. In place of a conditional probability on the index, we identify $p$ external variables $\mathbf{Z}$ referring to measurements, characteristics of the document d that directly imposes an effect on what topics will manifest in such document. Assuming these variables exist, the function for the per-document topic expressions is then revised into

\begin{equation}
    \label{eq:semipar2}
    P(t|d) \propto h(t|Z_{d1},Z_{d2},...,Z_{dp})
\end{equation}

\section{Simulation Study}
\label{sec:section4}

We discuss the design of the simulation study through which the performance of SemiparTM will be compared with those of the legacy methods across a range of simulation scenarios. Evaluation procedures and metrics are also outlined in this section.

The data generation process begins with defining the external variables that will determine topic distributions across documents:

\begin{equation}
    \label{eq:sim_dgp1}
    \begin{array}{ll}
    	Z_{1} & \sim Poisson(1) \\
    	Z_{2} & \sim Normal(20,7) \\
    	Z_{3} & \sim Bernoulli(0.8) \\
    	Z_{4} & \sim Beta(6,2) \\
    	Z_{5} & \sim Beta(10,2)
    \end{array}
\end{equation}

Assuming $T=10$ topics, the score $b_k$ of each document to the $k$th topic is determined by a hidden model $b_{0k}$ a function of $Z$ filtered by

\begin{equation}
    \label{eq:sim_dgp2}
    b_k = 
	\left\{
		\begin{array}{ll}
      		b_{0k} & \textrm{with probability } 1 - s \\
      		0 & \textrm{with probability } s \\
		\end{array} 
	\right.
\end{equation}

where $s$ is the data sparsity parameter, dictating what proportion of the topic and dictionary matrices are expected to be populated with zero cells. The topic models $b_{0k}$ will be specified differently, with the first four taking on linear forms (equation \ref{eq:topic1} thru \ref{eq:topic4}), the next three as nonlinear (\ref{eq:topic5}, \ref{eq:topic6}, \ref{eq:topic7}), and the remaining as having correlations with the other $b_k$ topic scores (\ref{eq:topic8}, \ref{eq:topic9}, \ref{eq:topic10}).

\begin{align}
b_{01} & = -1 + Z_1 + 0.2 Z_2 + Z_3 - 0.9Z_4 - 2 Z_5 + m\epsilon \label{eq:topic1} \\
b_{02} & = 3 + 1.5 Z_1 + 0.15 Z_2 - 5 Z_3 - 5 Z_5 + m\epsilon \label{eq:topic2} \\
b_{03} & = 2 + 0.2 Z_2 - 1.4 Z_1 + m\epsilon \label{eq:topic3} \\
b_{04} & = 1.6 Z_1 + 8 Z_3 - 9 Z_4 + m\epsilon \label{eq:topic4} \\
b_{05} & = \frac{Z_1^2}{5 Z_5} + m\epsilon \label{eq:topic5} \\
b_{06} & = 6 \sin(Z_5Z_1) + m\epsilon \label{eq:topic6} \\
b_{07} & = 2 + 3 Z_1 Z_4 - 2 Z_3 + m\epsilon \label{eq:topic7} \\
b_{08} & = 1 + 10 Z_4 - 2 b_3 + m\epsilon \label{eq:topic8} \\ 
b_{09} & = 0.2 Z_2 + 0.2 b_7 + m\epsilon \label{eq:topic9} \\
b_{0,10} & = -5 + 0.9 b_1 - 1.2 b_7+ m\epsilon \label{eq:topic10}
\end{align}

where $\epsilon \sim N(0,1)$. The value of $m$ will determine the level of misspecification error as it affect the model by inflating the variance of the error term thus contaminating model specification. In the simulations, $m$ will be set to have two values, $m \in  \{1,2\}$. Combining each $b$ will result in a matrix the simulated matrix $\mathbf{B}_{10 \times D} = [b^{'}_1,b^{'}_2,b^{'}_3, ..., b^{'}_{10}]^{'} $.

To simulate a sparse $\mathbf{X}_{W \times 10} = {x_kj}$ matrix according to the same data sparsity parameter $s$ used in (\ref{eq:sim_dgpx}), we generate random samples from a zero-inflated Poisson distribution as

\begin{equation}
    \label{eq:sim_dgpx}
    x_{kj} = \frac{1}{90} x^{*}_{kj}; x^{*}_{kj} \sim f(x^{*}_{kj}) = 
	\left\{
		\begin{array}{ll}
      		s + (1 - s)e^{-100} & x^{*}_{kj} = 0 \\
      		(1 - s)\frac{100 x^{*}_{kj} e^{-100}}{x^{*}_{kj}!} & x^{*}_{kj} > 0 \\
		\end{array} 
	\right.
\end{equation}

The process as presented here proceeds in a backwards direction, beginning with identifying the auxiliary information (the $Z_l$ variables in equation (\ref{eq:sim_dgp1})) and then connecting them to the topic scores (equations (\ref{eq:topic1}) to (\ref{eq:topic10})), which will then produce the corpus $\mathbf{Y}$. In practice, analysis begins with collecting the corpus $\mathbf{Y}$ and using topic modeling to discover the topic and dictionary matrices. In SemiparTM, the regression stage will then be used to uncover the connection between the topic scores in $\mathbf{B}$ with the $Z_l$ variables.

The analyst is tasked with identifying what variables are to be used for each $Z_l$. The distributions proposed in equation (\ref{eq:sim_dgp1}) represent only an example of what, returning to the restaurant company scenario presented in Section \ref{sec:section1}, may likely be the variables accessible to an analyst and that may determine the type of feedback the company will receive. The Poisson distribution assumed for $Z_1$ may represent the average number of times that a customer in the feedback database has sent a complaint to the company (on average, a customer may send only one complaint in their entire lifetime, but there exist certain customers that periodically send in feedback). Meanwhile, the normal distribution in $Z_2$ may represent the number of years that the restaurant or branch that received the complaint has been in operation (average of 20 years); $Z_3$ an indicator variable of whether that branch has more than 1 business channel (Dine In, Take Home, Delivery, Parties and Business Functions, Drive-Thru, etc.), while $Z_4$ and $Z_5$ are scores that the store received during a corporate audit.

These variables, in fact, represent actual variables that were used in the application of the proposed method to a corpus of customer feedback obtained from a food service establishment. The distributions and their parameters were set so as to approximate the possible distributions of these data (see histograms of actual data in Figure 
\ref{fig:docustats}). However, it should be noted that the proposed method is not limited to the above distributions, and these may be replaced (or supplemented) by other variables depending on availability and accessibility.

Simulations are performed across settings of the following parameters: corpus size, vocabulary size, number of underlying topics, and presence of topic correlations. These settings take into consideration known weaknesses of different methods noted in the literature.

\begin{table}
\caption{Levels used for each simulation parameter. Simulations are based on the combinations of each of these parameters, including the two levels of the misspecification error $m$. The designation of ``small", ``medium", and ``large" refer primarily to the size and dimensionality of the corpus. Meanwhile, sparsity levels on these three levels can be interpreted as being in the ``low" sparsity, ``medium" sparsity, and ``high" sparsity scenarios.}
\centering
\begin{tabular}{p{6.5cm} p{2cm} p{2cm} p{2cm}}
\hline
\textbf{Parameter} & \textbf{Small} & \textbf{Med} & \textbf{Large} \\ \hline
Document           & 150          & 1000         & 3000          \\
Words              & 500          & 1500         & 3500          \\
Sparsity           & 0.70         & 0.90         & 0.99          \\ \hline
\end{tabular}
\label{tab:sim_param}
\end{table}

Simulations are based on the combinations of each of the parameters in Table \ref{tab:sim_param}, including the two levels of the misspecification error ($m = 1,2$). The designation of ``small", ``medium", and ``large" refer primarily to the size and dimensionality of the corpus. Meanwhile, sparsity levels on these three levels can be interpreted as being in the ``low" sparsity, ``medium" sparsity, and ``high" sparsity scenarios. Simulation settings in Table \ref{tab:sim_param} have been designed to capture specific characteristics of the corpus of real feedback data obtained for the application presented in Section \ref{sec:section6}. We further discuss in Section \ref{sec:section6} further detail regarding this corpus and the results after applying SemiparTM and the legacy methods, but with a size of 253 documents and 844 unique words after cleaning, the corpus falls comfortably within the ``Small" category.

The number of underlying topics for this simulation study is fixed, i.e., it is assumed that actual underlying latent topics exist in the corpus, and that the analyst has prior knowledge of how many there are. Models are assessed based on how similar they are to the true topic distribution, the specifics of which will be discussed subsequently. This is to assess proposed claims towards instability of LDA by looking into how consistently and closely it reconstructs a set of true topic distributions. This would provide a much stronger argument for or against LDA, as prior studies cited have thus far focused only in replicating results. A simulation study comparing true versus estimated topic structures by LDA and the other topic models would allow for more objective comparisons between them.

Proximity between estimated and true topic structures will be evaluated through the cosine similarity measure. The algorithm runs as follows: given the true topic matrix $\mathbf{X}$ and its estimate $\hat{X}$ from one of the four methods, the sum in (\ref{eq:cosine1}) represents the total cosine similarity between the true and estimated matrix $\mathbf{X}$, in effect the total cosine similarity of a method's estimate to the true topic distribution. The same measure will be used in estimating the difference between the true and estimated topic distribution matrix $\mathbf{B}$, given in (\ref{eq:cosine2}).

\begin{equation} 
\label{eq:cosine1} 
S_C (\mathbf{X}, \hat{\mathbf{X}}) = \frac{1}{T} \sum_{k}{S_C(x_k, \hat{x}_k)} = \frac{1}{T} \sum_{k}{ \frac{x_k \cdot \hat{x}_k}{||x_k|| \times ||\hat{x}_k||}} 
\end{equation} 

\begin{equation} 
\label{eq:cosine2} 
S_C (\mathbf{B}, \hat{\mathbf{B}}) = \frac{1}{T} \sum_{k}{S_C(b_k, \hat{b}_k)} = \frac{1}{T} \sum_{k}{ \frac{b_k \cdot \hat{b}_k}{||b_k|| \times ||\hat{b}_k||}} 
\end{equation} 

\section{Results and Discussions}
\label{sec:section5}

Three versions of SemiparTM are implemented, each at varying configurations defined by the shrinkage penalty: SemiparTM-1 using a shrinkage penalty fixed at 1, SemiparTM-3 using a penalty fixed at 3,  and SemiparTM-cv using the cross-validation algorithm outlined in Section 3 to adaptively set the penalty. We compare the performances of the four methods on varying sizes (in number of documents) and dimensionality (number of unique words) of the corpus. Similarity scores are based on cosine similarity measures and characterizes the actual and predicted topic distribution.

Table \ref{tab:table2} exhibits superior similarity scores for the three configurations of SemiparTM over the legacy methods for the topic distribution matrix at least for the training corpus. PLSA is comparable to SemiparTM in fewer document-words combination. For the holdout corpus, similarity scores decline for all methods, but SemiparTM still manages to outperform the legacy methods for some document-word combinations of the corpora. In the training corpus, modeling of topic distributions by SemiparTM is accomplished by the nonnegative matrix factorization stage (Stage 1), while on the holdout corpus is done by the semiparametric regression stage (Stage 2). While similarity scores drop in the holdout corpus, performance can still be improved when the training corpus includes more documents, which in consequence means more documents for the regression stage to use in tuning the estimates. It is noted that this is equivalent to some decline in the cosine similarity for the training corpus. 

Increasing the size of the training corpus from 150 to 3000 led to a decrease in average similarity score from 0.690 to 0.584 in the cross-validated SemiparTM (assuming a fixed vocabulary of 500 words). However, on the holdout corpus, this was an equivalent to an increase from 0.104 to 0.450, which nearly closed the gap in average similarity score between the two corpora.

% ======== table 2 start
\begin{table}[!h]

\caption{\label{tab:table2}Average cosine similarities on the topic distribution and dictionary matrices, across levels of document and word count}
\centering
\begin{tabu} to \linewidth {>{\raggedright\arraybackslash}p{3cm}>{\raggedleft\arraybackslash}p{2cm}>{\raggedleft}X>{\raggedleft}X>{\raggedleft}X>{\raggedleft}X>{\raggedleft}X>{\raggedleft}X>{\raggedleft}X>{\raggedleft}X>{\raggedleft}X}
\toprule
\multicolumn{2}{c}{ } & \multicolumn{6}{c}{Topic Distribution Matrix} & \multicolumn{3}{c}{Dictionary Matrix} \\
\cmidrule(l{3pt}r{3pt}){3-8} \cmidrule(l{3pt}r{3pt}){9-11}
\multicolumn{2}{c}{ } & \multicolumn{3}{c}{Training} & \multicolumn{3}{c}{Holdout} & \multicolumn{3}{c}{Training} \\
\cmidrule(l{3pt}r{3pt}){3-5} \cmidrule(l{3pt}r{3pt}){6-8} \cmidrule(l{3pt}r{3pt}){9-11}
\multicolumn{2}{c}{ } & \multicolumn{3}{c}{\# Words} & \multicolumn{3}{c}{\# Words} & \multicolumn{3}{c}{\# Words} \\
\cmidrule(l{3pt}r{3pt}){3-5} \cmidrule(l{3pt}r{3pt}){6-8} \cmidrule(l{3pt}r{3pt}){9-11}
Method & \# Docs & 500 & 1500 & 3500 & 500 & 1500 & 3500 & 500 & 1500 & 3500\\
\midrule
 & 150 & 0.236 & 0.216 & 0.212 & 0.242 & 0.221 & 0.270 & 0.429 & 0.424 & 0.441\\

 & 1000 & 0.239 & 0.250 & 0.232 & 0.264 & 0.287 & 0.182 & 0.408 & 0.393 & 0.389\\

\multirow{-3}{3cm}{\raggedright\arraybackslash LSA} & 3000 & 0.236 & 0.330 & 0.242 & 0.234 & 0.338 & 0.231 & 0.398 & 0.438 & 0.384\\
\addlinespace
 & 150 & 0.770 & 0.732 & 0.806 & 0.102 & 0.134 & 0.169 & 0.618 & 0.572 & 0.632\\

 & 1000 & 0.631 & 0.631 & 0.582 & 0.088 & 0.109 & 0.053 & 0.318 & 0.318 & 0.283\\

\multirow{-3}{3cm}{\raggedright\arraybackslash PLSA} & 3000 & 0.661 & 0.602 & 0.702 & 0.083 & 0.082 & 0.086 & 0.358 & 0.312 & 0.404\\
\addlinespace
 & 150 & 0.224 & 0.182 & 0.171 & 0.234 & 0.174 & 0.211 & 0.417 & 0.375 & 0.392\\

 & 1000 & 0.234 & 0.230 & 0.219 & 0.255 & 0.290 & 0.174 & 0.309 & 0.298 & 0.321\\

\multirow{-3}{3cm}{\raggedright\arraybackslash LDA} & 3000 & 0.218 & 0.231 & 0.221 & 0.219 & 0.218 & 0.199 & 0.289 & 0.340 & 0.305\\
\addlinespace
 & 150 & 0.696 & 0.736 & 0.747 & 0.190 & 0.182 & 0.271 & 0.601 & 0.559 & 0.606\\

 & 1000 & 0.700 & 0.665 & 0.675 & 0.263 & 0.264 & 0.155 & 0.385 & 0.332 & 0.352\\

\multirow{-3}{3cm}{\raggedright\arraybackslash SemiparTM-1} & 3000 & 0.592 & 0.647 & 0.622 & 0.210 & 0.210 & 0.172 & 0.285 & 0.343 & 0.314\\
\addlinespace
 & 150 & 0.696 & 0.736 & 0.747 & 0.190 & 0.182 & 0.271 & 0.601 & 0.559 & 0.606\\

 & 1000 & 0.700 & 0.665 & 0.675 & 0.263 & 0.264 & 0.155 & 0.385 & 0.332 & 0.352\\

\multirow{-3}{3cm}{\raggedright\arraybackslash SemiparTM-3} & 3000 & 0.592 & 0.647 & 0.622 & 0.210 & 0.210 & 0.172 & 0.285 & 0.343 & 0.314\\
\addlinespace
 & 150 & 0.690 & 0.736 & 0.739 & 0.104 & 0.096 & 0.117 & 0.601 & 0.559 & 0.605\\

 & 1000 & 0.698 & 0.664 & 0.667 & 0.295 & 0.254 & 0.181 & 0.384 & 0.331 & 0.350\\

\multirow{-3}{3cm}{\raggedright\arraybackslash SemiparTM-cv} & 3000 & 0.584 & 0.626 & 0.622 & 0.450 & 0.410 & 0.386 & 0.276 & 0.332 & 0.309\\
\bottomrule
\end{tabu}
\end{table}
% ======== table 2 end

% ======== table 3 start
\begin{table}[!h]

\caption{\label{tab:table3}Average cosine similarities on the topic distribution and dictionary matrices, across levels of document count and sparsity level}
\centering
\begin{tabu} to \linewidth {>{\raggedright\arraybackslash}p{3cm}>{\raggedleft\arraybackslash}p{2cm}>{\raggedleft}X>{\raggedleft}X>{\raggedleft}X>{\raggedleft}X>{\raggedleft}X>{\raggedleft}X>{\raggedleft}X>{\raggedleft}X>{\raggedleft}X}
\toprule
\multicolumn{2}{c}{ } & \multicolumn{6}{c}{Topic Distribution Matrix} & \multicolumn{3}{c}{Dictionary Matrix} \\
\cmidrule(l{3pt}r{3pt}){3-8} \cmidrule(l{3pt}r{3pt}){9-11}
\multicolumn{2}{c}{ } & \multicolumn{3}{c}{Training} & \multicolumn{3}{c}{Holdout} & \multicolumn{3}{c}{Training} \\
\cmidrule(l{3pt}r{3pt}){3-5} \cmidrule(l{3pt}r{3pt}){6-8} \cmidrule(l{3pt}r{3pt}){9-11}
\multicolumn{2}{c}{ } & \multicolumn{3}{c}{Sparsity} & \multicolumn{3}{c}{Sparsity} & \multicolumn{3}{c}{Sparsity} \\
\cmidrule(l{3pt}r{3pt}){3-5} \cmidrule(l{3pt}r{3pt}){6-8} \cmidrule(l{3pt}r{3pt}){9-11}
Method & \# Docs & 0.70 & 0.90 & 0.99 & 0.70 & 0.90 & 0.99 & 0.70 & 0.90 & 0.99\\
\midrule
 & 150 & 0.382 & 0.221 & 0.061 & 0.426 & 0.259 & 0.048 & 0.555 & 0.415 & 0.323\\

 & 1000 & 0.418 & 0.233 & 0.070 & 0.415 & 0.243 & 0.074 & 0.525 & 0.356 & 0.309\\

\multirow{-3}{3cm}{\raggedright\arraybackslash LSA} & 3000 & 0.415 & 0.236 & 0.062 & 0.423 & 0.259 & 0.018 & 0.512 & 0.364 & 0.312\\
\addlinespace
 & 150 & 0.775 & 0.733 & 0.800 & 0.250 & 0.131 & 0.025 & 0.845 & 0.671 & 0.307\\

 & 1000 & 0.676 & 0.588 & 0.580 & 0.148 & 0.084 & 0.018 & 0.520 & 0.269 & 0.129\\

\multirow{-3}{3cm}{\raggedright\arraybackslash PLSA} & 3000 & 0.740 & 0.616 & 0.620 & 0.168 & 0.078 & 0.004 & 0.528 & 0.281 & 0.264\\
\addlinespace
 & 150 & 0.323 & 0.189 & 0.065 & 0.337 & 0.215 & 0.067 & 0.582 & 0.398 & 0.204\\

 & 1000 & 0.377 & 0.234 & 0.073 & 0.392 & 0.244 & 0.083 & 0.455 & 0.281 & 0.192\\

\multirow{-3}{3cm}{\raggedright\arraybackslash LDA} & 3000 & 0.368 & 0.229 & 0.064 & 0.388 & 0.247 & 0.012 & 0.463 & 0.250 & 0.182\\
\addlinespace
 & 150 & 0.774 & 0.682 & 0.721 & 0.382 & 0.173 & 0.088 & 0.840 & 0.667 & 0.258\\

 & 1000 & 0.722 & 0.647 & 0.672 & 0.381 & 0.234 & 0.067 & 0.505 & 0.332 & 0.232\\

\multirow{-3}{3cm}{\raggedright\arraybackslash SemiparTM-1} & 3000 & 0.695 & 0.585 & 0.532 & 0.393 & 0.218 & 0.003 & 0.500 & 0.230 & 0.163\\
\addlinespace
 & 150 & 0.774 & 0.682 & 0.721 & 0.382 & 0.173 & 0.088 & 0.840 & 0.667 & 0.258\\

 & 1000 & 0.722 & 0.647 & 0.672 & 0.381 & 0.234 & 0.067 & 0.505 & 0.332 & 0.232\\

\multirow{-3}{3cm}{\raggedright\arraybackslash SemiparTM-3} & 3000 & 0.695 & 0.585 & 0.532 & 0.393 & 0.218 & 0.003 & 0.500 & 0.230 & 0.163\\
\addlinespace
 & 150 & 0.762 & 0.682 & 0.721 & 0.249 & 0.065 & 0.001 & 0.839 & 0.667 & 0.258\\

 & 1000 & 0.712 & 0.646 & 0.672 & 0.594 & 0.135 & 0.001 & 0.502 & 0.331 & 0.232\\

\multirow{-3}{3cm}{\raggedright\arraybackslash SemiparTM-cv} & 3000 & 0.681 & 0.574 & 0.531 & 0.995 & 0.310 & 0.000 & 0.489 & 0.217 & 0.161\\
\bottomrule
\end{tabu}
\end{table}
% ======== table 3 end

% ======== table 4 start
\begin{table}[!h]

\caption{\label{tab:table4}Average cosine similarities on the topic distribution and dictionary matrices, across levels of document count and misspecification error}
\centering
\begin{tabu} to \linewidth {>{\raggedright\arraybackslash}p{3cm}>{\raggedleft\arraybackslash}p{2cm}>{\raggedleft}X>{\raggedleft}X>{\raggedleft}X>{\raggedleft}X>{\raggedleft}X>{\raggedleft}X}
\toprule
\multicolumn{2}{c}{ } & \multicolumn{4}{c}{Topic Distribution Matrix} & \multicolumn{2}{c}{Dictionary Matrix} \\
\cmidrule(l{3pt}r{3pt}){3-6} \cmidrule(l{3pt}r{3pt}){7-8}
\multicolumn{2}{c}{ } & \multicolumn{2}{c}{Training} & \multicolumn{2}{c}{Holdout} & \multicolumn{2}{c}{Training} \\
\cmidrule(l{3pt}r{3pt}){3-4} \cmidrule(l{3pt}r{3pt}){5-6} \cmidrule(l{3pt}r{3pt}){7-8}
\multicolumn{2}{c}{ } & \multicolumn{2}{c}{Misspecification Error} & \multicolumn{2}{c}{Misspecification Error} & \multicolumn{2}{c}{Misspecification Error} \\
\cmidrule(l{3pt}r{3pt}){3-4} \cmidrule(l{3pt}r{3pt}){5-6} \cmidrule(l{3pt}r{3pt}){7-8}
Method & \# Docs & None & Present & None & Present & None & Present\\
\midrule
 & 150 & 0.228 & 0.215 & 0.247 & 0.241 & 0.431 & 0.432\\

 & 1000 & 0.248 & 0.233 & 0.249 & 0.239 & 0.398 & 0.395\\

\multirow{-3}{3cm}{\raggedright\arraybackslash LSA} & 3000 & 0.251 & 0.242 & 0.253 & 0.236 & 0.390 & 0.410\\
\addlinespace
 & 150 & 0.786 & 0.753 & 0.136 & 0.134 & 0.612 & 0.603\\

 & 1000 & 0.611 & 0.618 & 0.089 & 0.077 & 0.316 & 0.296\\

\multirow{-3}{3cm}{\raggedright\arraybackslash PLSA} & 3000 & 0.652 & 0.666 & 0.086 & 0.080 & 0.359 & 0.357\\
\addlinespace
 & 150 & 0.211 & 0.174 & 0.225 & 0.187 & 0.396 & 0.393\\

 & 1000 & 0.239 & 0.217 & 0.248 & 0.231 & 0.306 & 0.313\\

\multirow{-3}{3cm}{\raggedright\arraybackslash LDA} & 3000 & 0.225 & 0.215 & 0.224 & 0.208 & 0.301 & 0.296\\
\addlinespace
 & 150 & 0.749 & 0.703 & 0.220 & 0.209 & 0.589 & 0.588\\

 & 1000 & 0.680 & 0.680 & 0.236 & 0.218 & 0.365 & 0.347\\

\multirow{-3}{3cm}{\raggedright\arraybackslash SemiparTM-1} & 3000 & 0.592 & 0.616 & 0.210 & 0.199 & 0.289 & 0.305\\
\addlinespace
 & 150 & 0.749 & 0.703 & 0.220 & 0.209 & 0.589 & 0.588\\

 & 1000 & 0.680 & 0.680 & 0.236 & 0.218 & 0.365 & 0.347\\

\multirow{-3}{3cm}{\raggedright\arraybackslash SemiparTM-3} & 3000 & 0.592 & 0.616 & 0.210 & 0.199 & 0.289 & 0.305\\
\addlinespace
 & 150 & 0.746 & 0.698 & 0.112 & 0.099 & 0.589 & 0.587\\

 & 1000 & 0.675 & 0.678 & 0.279 & 0.208 & 0.365 & 0.344\\

\multirow{-3}{3cm}{\raggedright\arraybackslash SemiparTM-cv} & 3000 & 0.586 & 0.604 & 0.458 & 0.412 & 0.280 & 0.298\\
\bottomrule
\end{tabu}
\end{table}
% ======== table 4 end

Similar patterns can be observed in the similarity scores for the estimated dictionary matrices, although here there is no comparison to be made between training and holdout corpus as the dictionary is held fixed for the two corpora. Increasing the number of documents has the effect of diminishing quality of estimation in terms of their cosine similarities. Although in both topic distribution and dictionary matrices, the decrease in similarity scores when increasing the number of documents from 1000 to 3000 appears to be minimal in comparison to changes from 150 to 1000.

The behavior of SemiparTM across levels of sparsity in the corpus is compared to the legacy methods for both the training and holdout corpora in Table \ref{tab:table3}. With regards to performance on the topic distribution matrix, however, this shows a narrative that has already been explored in Table \ref{tab:table2}, with SemiparTM configurations demonstrating superior performance compared with the legacy methods on the training corpus in terms of how well they estimated the topic distribution matrix. Only here, PLSA yield more comparable similarity measures to SemiparTM. Increasing the level of sparsity doesn't put much of a burden on either PLSA or the SemiparTM configurations on the training corpus. Again, it is on the holdout corpus where this narrative changes. On the holdout corpus, SemiparTM not only drops in performance, but also appears to be extremely sensitive to increasing levels of sparsity. SemiparTM-cv, especially, nearly zeroes out with a corpus size of 3000 documents but with 0.99 sparsity. The other two SemiparTM configurations fare significantly better and even remain at comparable levels with the legacy methods, which at this point is admittedly low for all the methods.

Once more, no comparison is made with the holdout corpus for the dictionary matrix because these values are retained between training and holdout corpora. The SemiparTM configurations, along with PLSA, yielded higher cosine similarity scores than either LSA or PLSA in small corpuz sizes with low to medium sparsity (0.70 to 0.99), but with bigger corpora or higher levels of sparsity, this difference is quickly closed.

This next set of results analyzes the impact of misspecification errors on the methods. It has been observed before that SemiparTM, unlike the legacy methods, can take advantage of increasing number of documents in the training corpus as leverage for improving estimation performance on the holdout corpus. Thus, in Table \ref{tab:table4} despite a drop in cosine similarity between the two corpora, a drop that is also observed for PLSA, which yielded high similarity scores in the training corpus, SemiparTM is observed to actually increase in similarity scores as the size of the training corpus increases. This is a primary benefit of the semiparametric regression stage.

Even with misspecification error on the true models determining the relationship between each topic to the auxiliary information, SemiparTM configurations do not exhibit large declines in its similarity scores on the hold-out topic distribution matrix. In the cross-validated configuration of SemiparTM, the highest similarity score received in the large corpus in the presence of misspecification error ($m = 2$) is only a few units away in the equivalent setting without misspecification error. The nonparametric nature of the topic models used in SemiparTM contributes to its robustness to misclassification error. Thus, when the analyst is confounded with limited options on auxiliary variables to include, SemiparTM resolves this easily through its robustness to misspecification errors. 

Even the nonnegative matrix factorization step, which produced the dictionary matrix being assessed, handled misspecification errors quite well. At least in the small corpus setting, the SemiparTM configurations were able to consistently meet with PLSA at having the highest similarity scores under both levels of misspecification error.

%===Table 5 start
\begin{table}[!h]

\caption{\label{tab:table5}Average cosine similarities on the topic distribution and dictionary matrices, across levels of vocabulary size and misspecification error}
\centering
\begin{tabu} to \linewidth {>{\raggedright\arraybackslash}p{3cm}>{\raggedleft\arraybackslash}p{2cm}>{\raggedleft}X>{\raggedleft}X>{\raggedleft}X>{\raggedleft}X>{\raggedleft}X>{\raggedleft}X}
\toprule
\multicolumn{2}{c}{ } & \multicolumn{4}{c}{Topic Distribution Matrix} & \multicolumn{2}{c}{Dictionary Matrix} \\
\cmidrule(l{3pt}r{3pt}){3-6} \cmidrule(l{3pt}r{3pt}){7-8}
\multicolumn{2}{c}{ } & \multicolumn{2}{c}{Training} & \multicolumn{2}{c}{Holdout} & \multicolumn{2}{c}{Training} \\
\cmidrule(l{3pt}r{3pt}){3-4} \cmidrule(l{3pt}r{3pt}){5-6} \cmidrule(l{3pt}r{3pt}){7-8}
\multicolumn{2}{c}{ } & \multicolumn{2}{c}{Misspecification Error} & \multicolumn{2}{c}{Misspecification Error} & \multicolumn{2}{c}{Misspecification Error} \\
\cmidrule(l{3pt}r{3pt}){3-4} \cmidrule(l{3pt}r{3pt}){5-6} \cmidrule(l{3pt}r{3pt}){7-8}
Method & \# Words & None & Present & None & Present & None & Present\\
\midrule
 & 500 & 0.244 & 0.231 & 0.252 & 0.242 & 0.402 & 0.422\\

 & 1500 & 0.246 & 0.233 & 0.259 & 0.259 & 0.415 & 0.406\\

\multirow{-3}{3cm}{\raggedright\arraybackslash LSA} & 3500 & 0.231 & 0.217 & 0.234 & 0.218 & 0.416 & 0.409\\
\addlinespace
 & 500 & 0.690 & 0.685 & 0.096 & 0.086 & 0.441 & 0.422\\

 & 1500 & 0.681 & 0.667 & 0.116 & 0.120 & 0.439 & 0.427\\

\multirow{-3}{3cm}{\raggedright\arraybackslash PLSA} & 3500 & 0.697 & 0.692 & 0.114 & 0.103 & 0.455 & 0.450\\
\addlinespace
 & 500 & 0.236 & 0.214 & 0.247 & 0.225 & 0.345 & 0.332\\

 & 1500 & 0.221 & 0.195 & 0.241 & 0.220 & 0.326 & 0.347\\

\multirow{-3}{3cm}{\raggedright\arraybackslash LDA} & 3500 & 0.213 & 0.182 & 0.210 & 0.176 & 0.352 & 0.351\\
\addlinespace
 & 500 & 0.656 & 0.668 & 0.229 & 0.213 & 0.422 & 0.426\\

 & 1500 & 0.717 & 0.674 & 0.226 & 0.218 & 0.439 & 0.433\\

\multirow{-3}{3cm}{\raggedright\arraybackslash SemiparTM-1} & 3500 & 0.714 & 0.692 & 0.218 & 0.201 & 0.472 & 0.455\\
\addlinespace
 & 500 & 0.656 & 0.668 & 0.229 & 0.213 & 0.422 & 0.426\\

 & 1500 & 0.717 & 0.674 & 0.226 & 0.218 & 0.439 & 0.433\\

\multirow{-3}{3cm}{\raggedright\arraybackslash SemiparTM-3} & 3500 & 0.714 & 0.692 & 0.218 & 0.201 & 0.472 & 0.455\\
\addlinespace
 & 500 & 0.650 & 0.665 & 0.317 & 0.249 & 0.416 & 0.424\\

 & 1500 & 0.715 & 0.671 & 0.206 & 0.186 & 0.439 & 0.430\\

\multirow{-3}{3cm}{\raggedright\arraybackslash SemiparTM-cv} & 3500 & 0.708 & 0.683 & 0.186 & 0.155 & 0.473 & 0.452\\
\bottomrule
\end{tabu}
\end{table}
%===Table 5 end

Similar observations can be made in Table \ref{tab:table5}. Only at increased levels of dimensionality on the vocabulary the second stage of the method does see decreases in overall cosine similarities in the topic distribution matrix. However, this decreases still put the SemiparTM configurations at levels well comparable with the legacy methods, especially LDA. In fact, in the small corpus setting, SemiparTM-cv was still able to yield the highest similarity scores even with the presence of misspecification error.

Table \ref{tab:table6} explores cosine similarities across joint impacts of both sparsity and misspecification. It has been noted that SemiparTM is sensitive to sparsity, but is protected against significant performance declines due to misspecification in the regression models, and this has been visualized exactly in both tables. For both the training and the holdout corpus, while the cosine similarity may be declining from top to bottom (increasing sparsity), the differences tend to be small with or without misspecification error. That the pattern of the values tend to be the same for all methods suggests that across levels of sparsity and misspecification error, SemiparTM in general performs at comparably the same levels as the legacy methods. In terms of cosine similarity, it is only in terms of dimensionality and corpus size at which this general performance really begins to be differentiated among them. In particular, it has been observed numerous times through these simulation results that SemiparTM tends to be better-suited at modeling small corpora.

% === Table 6 start
\begin{table}[!h]

\caption{\label{tab:table6}Average cosine similarities on the topic distribution and dictionary matrices, across levels of sparsity and misspecification error}
\centering
\begin{tabu} to \linewidth {>{\raggedright\arraybackslash}p{3cm}>{\raggedleft\arraybackslash}p{2cm}>{\raggedleft}X>{\raggedleft}X>{\raggedleft}X>{\raggedleft}X>{\raggedleft}X>{\raggedleft}X}
\toprule
\multicolumn{2}{c}{ } & \multicolumn{4}{c}{Topic Distribution Matrix} & \multicolumn{2}{c}{Dictionary Matrix} \\
\cmidrule(l{3pt}r{3pt}){3-6} \cmidrule(l{3pt}r{3pt}){7-8}
\multicolumn{2}{c}{ } & \multicolumn{2}{c}{Training} & \multicolumn{2}{c}{Holdout} & \multicolumn{2}{c}{Training} \\
\cmidrule(l{3pt}r{3pt}){3-4} \cmidrule(l{3pt}r{3pt}){5-6} \cmidrule(l{3pt}r{3pt}){7-8}
\multicolumn{2}{c}{ } & \multicolumn{2}{c}{Misspecification Error} & \multicolumn{2}{c}{Misspecification Error} & \multicolumn{2}{c}{Misspecification Error} \\
\cmidrule(l{3pt}r{3pt}){3-4} \cmidrule(l{3pt}r{3pt}){5-6} \cmidrule(l{3pt}r{3pt}){7-8}
Method & Spar & None & Present & None & Present & None & Present\\
\midrule
 & 0.70 & 0.411 & 0.395 & 0.431 & 0.411 & 0.532 & 0.537\\

 & 0.90 & 0.238 & 0.220 & 0.255 & 0.250 & 0.382 & 0.381\\

\multirow{-3}{3cm}{\raggedright\arraybackslash LSA} & 0.99 & 0.068 & 0.062 & 0.054 & 0.054 & 0.313 & 0.318\\
\addlinespace
 & 0.70 & 0.726 & 0.730 & 0.197 & 0.189 & 0.658 & 0.649\\

 & 0.90 & 0.648 & 0.656 & 0.108 & 0.097 & 0.448 & 0.421\\

\multirow{-3}{3cm}{\raggedright\arraybackslash PLSA} & 0.99 & 0.694 & 0.659 & 0.018 & 0.018 & 0.227 & 0.226\\
\addlinespace
 & 0.70 & 0.385 & 0.322 & 0.404 & 0.334 & 0.515 & 0.501\\

 & 0.90 & 0.217 & 0.212 & 0.233 & 0.232 & 0.318 & 0.327\\

\multirow{-3}{3cm}{\raggedright\arraybackslash LDA} & 0.99 & 0.073 & 0.063 & 0.066 & 0.060 & 0.192 & 0.199\\
\addlinespace
 & 0.70 & 0.740 & 0.736 & 0.397 & 0.371 & 0.645 & 0.635\\

 & 0.90 & 0.651 & 0.647 & 0.214 & 0.199 & 0.444 & 0.453\\

\multirow{-3}{3cm}{\raggedright\arraybackslash SemiparTM-1} & 0.99 & 0.683 & 0.648 & 0.064 & 0.063 & 0.236 & 0.223\\
\addlinespace
 & 0.70 & 0.740 & 0.736 & 0.397 & 0.371 & 0.645 & 0.635\\

 & 0.90 & 0.651 & 0.647 & 0.214 & 0.199 & 0.444 & 0.453\\

\multirow{-3}{3cm}{\raggedright\arraybackslash SemiparTM-3} & 0.99 & 0.683 & 0.648 & 0.064 & 0.063 & 0.236 & 0.223\\
\addlinespace
 & 0.70 & 0.730 & 0.722 & 0.572 & 0.488 & 0.642 & 0.630\\

 & 0.90 & 0.648 & 0.647 & 0.162 & 0.118 & 0.441 & 0.450\\

\multirow{-3}{3cm}{\raggedright\arraybackslash SemiparTM-cv} & 0.99 & 0.683 & 0.647 & 0.001 & 0.001 & 0.236 & 0.222\\
\bottomrule
\end{tabu}
\end{table}
% === Table 6 end

\section{Application to actual customer feedback corpus}
\label{sec:section6}

We demonstrate the use of SemiparTM on an actual corpus of customer feedback data obtained from a food service establishment. After cleaning, the corpus contained a set of 253 documents with a corpus of 844 unique words. Document length is an average of 19.9 words. Corpus sparsity is at 0.79.

\begin{figure}
	\centering
	\includegraphics[scale=0.25]{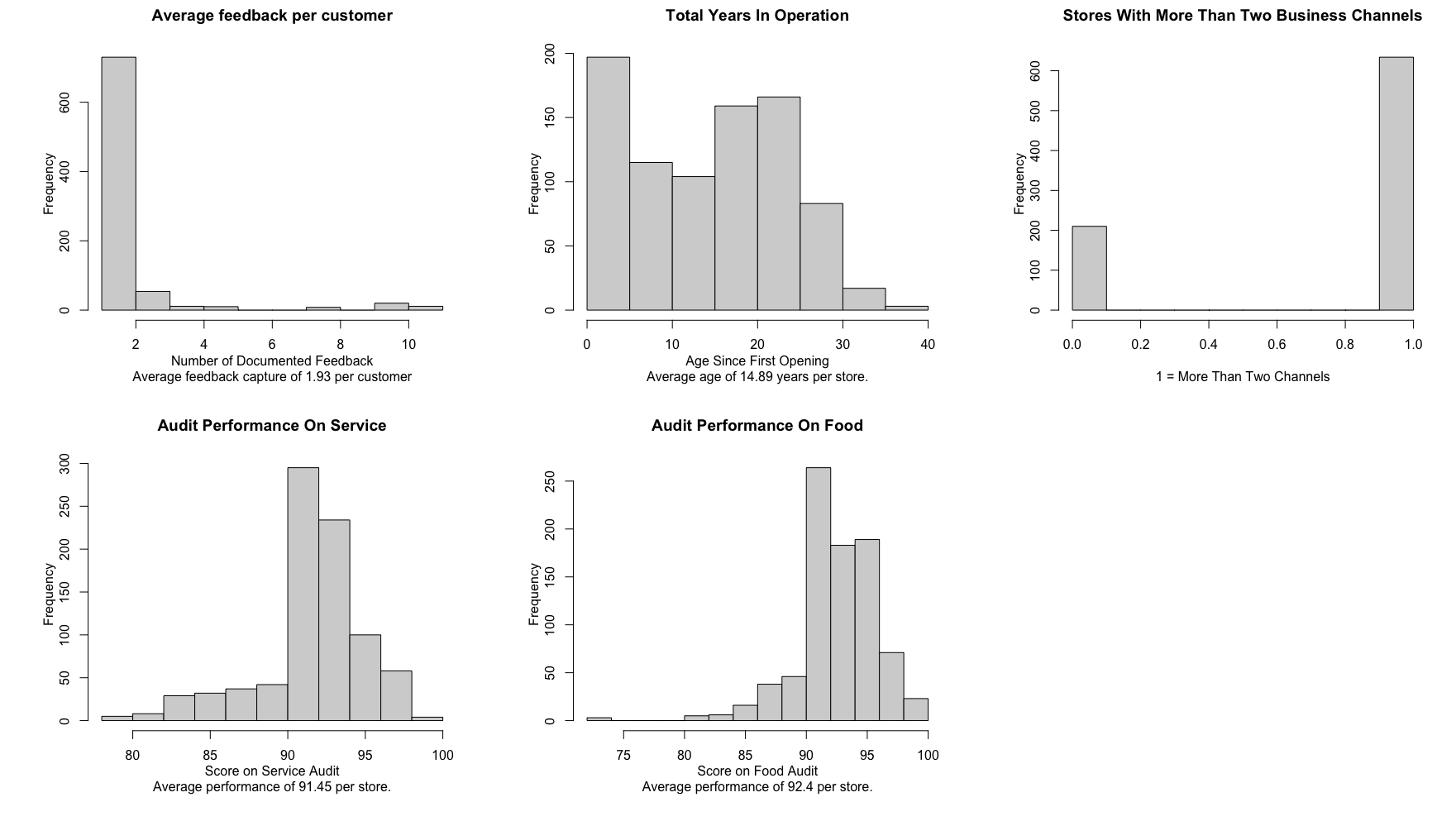}
	\caption{Histogram of auxiliary variables used in modeling SemiparTM on real consumer feedback.}
	\label{fig:docustats}
\end{figure}

For the semiparametric topic model, auxiliary variables used are the following: (1) average number of feedback received from the authors of each document (customers sending in the feedback), (2) length of operations in number of years since first opening, (3) a binary variable on whether or not the store being referenced in the feedback services more than two business channels, (4) average performance in a service audit, and (5) average performance in a product quality audit. Histograms of these auxiliary variables are presented in Figure \ref{fig:docustats}. Table \ref{tab:table7} presents the first four topics produced by LSA, PLSA, LDA, and SemiparTM-cv under the assumption of 10 latent topics in the corpus.

One important observation to take note of here is that while the simulation study guarantees that these topic definitions, coming from the dictionary matrix, are expected to be closer to the true topic distributions in this corpus size and sparsity level, interpretability is another matter altogether. For topic 1, SemiparTM-cv associated words that have much to do with payments and transactions: words like ``redeem", ``offlin" and specific bank names indicate possible issues that the business's customers might be having with regards to transacting for products and services. For LDA, this topic covers some overlapping words (``onlin", ``card", and ``payment"). For LSA and PLSA, this topic seems to be more general towards service-related issues, with words like ``manag", ``service", ``cashier", and ``time".

It is typical when performing topic modeling that these topics are interpreted based on their top associated words, using which they are given a name or some identifying description. Topic 1 for SemiparTM-cv may then be on payment and redemption issues,  while topic 2 covers promo-related issues (``branch", ``promo", ``free" and mention of product names).

While these topic definitions appear to be comparably useful between the methods, it is nevertheless noted that results of the simulation study warn that for LSA and LDA, quality of these word associations (at least according to the dictionary's cosine similarity) may be suspect in this particular case when the corpora is particularly small and covering a similarly small vocabulary.

% === Table 7 start
\begin{table}[!h]

\caption{\label{tab:table7}First four topics from SemiparTM-cv and the legacy methods. Note: The table presents the first four topics from SemiparTM-cv and the legacy methods as defined by their corresponding top 20 words with the highest scores in the dictionary matrix X. Sensitive information, such as tokens pertaining to brand names, product names, and the likes have been hidden with placeholder tokens.}
\centering
\fontsize{8}{9}\selectfont
\begin{tabu} to \linewidth {>{\raggedleft}X>{\raggedright}X>{\raggedright}X>{\raggedright}X>{\raggedright}X>{\raggedright}X>{\raggedright}X>{\raggedright}X>{\raggedright}X}
\toprule
\multicolumn{1}{c}{ } & \multicolumn{8}{c}{Method} \\
\cmidrule(l{3pt}r{3pt}){2-9}
\multicolumn{1}{c}{Topic} & \multicolumn{2}{c}{LSA} & \multicolumn{2}{c}{PLSA} & \multicolumn{2}{c}{LDA} & \multicolumn{2}{c}{SemiparTM-cv} \\
\cmidrule(l{3pt}r{3pt}){1-1} \cmidrule(l{3pt}r{3pt}){2-3} \cmidrule(l{3pt}r{3pt}){4-5} \cmidrule(l{3pt}r{3pt}){6-7} \cmidrule(l{3pt}r{3pt}){8-9}
 1 & sour & servic & deliveri & manag & [product name] & onlin & [brand] & hindi\\
 & sweet & manag & payment & branch & [product name] & card & redeem & sabi\\

 & spici & time & deliv & time & receiv & pleas & treat & wala\\

 & dark & food & call & crew & meal & alreadi & alway & sana\\

 & implement & store & card & servic & [product name] & refund & offlin & ano\\

 & measur & wait & onlin & cashier & [bank] & payment & becaus & bigay\\

 & salamat & cashier & cancel & becaus & [product name] & transact & [bank] & kayo\\

 & asap & branch & refund & receipt & [product name] & receiv & mall & tapo\\

 & takeout & becaus & fail & follow & text & cancel & complaint & walang\\

\multirow{-9}{*}{\raggedleft\arraybackslash} & deduct & crew & transact & guest & pcs & attach & food & nalang\\
\cmidrule{1-9}
 2 & sobra & alreadi & credit & guy & qualiti & email & branch & ulit\\

 & balik & hour & email & expect & correct & paid & card & ngayon\\

 & account & payment & pleas & wait & pc & fail & onlin & branch\\

 & lack & follow & alreadi & happen & oil & regard & [product name] & kain\\

 & foodpanda & deliveri & receiv & counter & takeout & famili & day & crew\\

 & hintay & guy & paid & [brand] & immedi & charg & [location] & tawag\\

 & [location] & card & pm & minut & pm & credit & promo & sama\\

 & bayad & veri & deduct & told & valu & pay & onli & kulang\\

 & happi & refund & contact & onli & [product name] & refer & experi & deliv\\

\multirow{-10}{*}{\raggedleft\arraybackslash} & oil & [brand] & account & sabi & measur & subject & free & bakit\\
\cmidrule{1-9}
 3 & [brand] & bag & [product name] & [brand] & fri & hindi & food & custom\\

 & food & plastic & [product name] & redeem & kulang & wala & branch & store\\

 & hindi & hindi & [product name] & treat & [product name] & sabi & time & call\\

 & branch & [product name] & receiv & alway & drink & sana & cashier & feedback\\

 & [product name] & [product name] & meal & offlin & [product name] & bigay & manag & rider\\

 & becaus & [product name] & [product name] & becaus & check & walang & servic & avail\\

 & treat & [product name] & qualiti & [bank] & [product name] & kayo & crew & offer\\

 & redeem & call & [product name] & branch & larg & tawag & becaus & befor\\

 & sabi & wast & takeout & mall & [product name] & ngayon & serv & accept\\

\multirow{-10}{*}{\raggedleft\arraybackslash} & wala & wala & oil & complaint & concern & tapo & receipt & pleas\\
\cmidrule{1-9}
 4 & time & deliveri & measur & [product name] & [product name] & ano & onli & follow\\

 & offlin & redeem & pm & card & bag & kanina & told & ms\\

 & [product name] & treat & dark & onli & plastic & lagay & sinc & gc\\

 & alway & sabi & implement & onlin & miss & isang & wait & late\\

 & pleas & [brand] & correct & promo & [product name] & nalang & store & wait\\

 & onli & fri & pcs & day & item & kain & hope & note\\

 & cashier & deliv & valu & [location] & regular & dumat & veri & regard\\

 & day & offlin & text & experi & [product name] & sobra & disappoint & inform\\

 & manag & concern & fri & veri & ice & sama & whi & tawag\\

\multirow{-10}{*}{\raggedleft\arraybackslash} & veri & drink & immedi & free & [product name] & bakit & [brand] & bigay\\
\bottomrule
\end{tabu}
\end{table}
\clearpage
% === Table 7 end

\section{Conclusions}
\label{sec:section7}

The proposed semiparametric topic modeling methodology (SemiparTM) combines known advantages of Latent Semantic Analysis (LSA) for discovering hidden semantic structures or topics in a collection of documents, and the predictive power of semiparametric regression techniques for generating topic distributions of previously unseen documents using a set of auxiliary document information. It has been demonstrated that SemiparTM performs at least at par or better than the three legacy topic models when reconstructing latent topic and vocabulary distributions. This advantage is best observed in small corpora with limited vocabularies.

The use of regression as a second stage of the model proved its benefits when applied to a new set of documents. Unlike PLSA, whose estimation tends to overfit to the training corpus and thus demonstrated significantly worse performance when given new documents, SemiparTM continued to perform at least at the level of LDA. Moreover, when the penalty parameter in the nonnegative LSA stage is tuned via cross-validation, SemiparTM exhibits a definitively superior performance in predicting the topic distributions of documents in a holdout corpus.

SemiparTM was also able to match PLSA with superior cosine similarities in the training corpora even with increasing levels of sparsity. The impact of sparsity is stronger when switching to holdout corpora, and it is here that we observe the second stage's sensitivity to levels of sparsity. Nevertheless, for small training corpora (150 documents), SemiparTM was still observed to be capable of performing better than PLSA and at par with LDA even at the 0.99 sparsity level. For simply modeling topic structures through the dictionary matrix, the SemiparTM configurations, along with PLSA, yielded higher cosine similarity scores than either LSA or LDA in small corpus sizes with low to medium sparsity (0.70 to 0.90).

Even in the presence of misspecification errors, SemiparTM configurations did not exhibit any large declines in similarity scores even in the hold-out topic matrix. In fact, the cross-validated SemiparTM achieved similarity scores under misspecification that are only a few units off from the same scores under no misspecification.

This study presents SemiparTM as a viable alternative to existing topic modeling procedures, one that accounts for additional auxiliary information without increasing the overall complexity of the model, and allows for easier automation through its reduced dependence on hyperparameters that would require prior tuning. The following chapter now closes our discussion by providing some practical recommendations for applying SemiparTM for real data analytic projects.

\section*{Acknowledgments}
\label{sec:section8}

Computations performed for this paper was made possible with the support of the Advanced Science and Technology Institute of the Department of Science and Technology (DOST-ASTI), which generously granted us access to their high performance computing facilities of their Computing and Archiving Environment (COARE).

\bibliographystyle{unsrtnat}
\bibliography{references} 

\end{document}